# Performance and Practical Considerations of Large and Small Language Models in Clinical Decision Support in Rheumatology

Sabine Felde[1], Rüdiger Buchkremer[1], Gamal Chehab[2,3,5], Christian Thielscher[4], Jörg HW Distler[2,3] Matthias Schneider[2,3] Jutta G. Richter [2,3]

---

[1] Institute of IT Management and Digitization Research (IFID), FOM University of Applied Sciences, 40476 Dusseldorf, Germany. [2] Department of Rheumatology, University Hospital Düsseldorf, Heinrich-Heine-University, Düsseldorf, Germany. [3] Hiller Research Center, Heinrich-Heine-University, Düsseldorf, Germany [4] FOM University of Applied Sciences, Competence Center for Medical Economics, Essen, Germany. [5] Department of Rheumatology and Clinical Immunology, KEM Kliniken Essen-Mitte, Essen, Germany.

**Large language models (LLMs) show promise for supporting clinical decision-making in complex fields such as rheumatology. Our evaluation shows that smaller language models (SLMs), combined with retrieval-augmented generation (RAG), achieve higher diagnostic and therapeutic performance than larger models, while requiring substantially less energy and enabling cost-efficient, local deployment. These features are attractive for resource-limited healthcare. However, expert oversight remains essential, as no model consistently reached specialist-level accuracy in rheumatology.**

Artificial intelligence (AI), particularly the application of large language models (LLMs) such as the prominent GPT-4o, offers emerging opportunities for optimizing clinical workflows, especially in diagnostics and therapeutic decision-making[1,2]. Language models (LMs) are capable of processing extensive textual data, extracting patterns, and generating contextually relevant, evidence-based treatment recommendations, thereby promoting consistency, quality, and standardization in clinical care[1,2]. The clinical benefit of LLM implementation in care is evident, but infrastructure demands associated with operating large-scale LLMs are considerable — particularly for hospitals with constrained IT budgets, limited energy capacity, scarce technical staff, and limited medical staff already loaded with administrative obligations - and the economic burden is high[3]. In such settings, smaller, locally deployable models (SLMs) offer a more cost-effective and operationally viable alternative. At present we define SLMs as those with fewer than 100 billion parameters, in contrast to frontier LLMs, which typically exceed several hundred billion and often reach into the trillions.

Rheumatology, a field characterized by high symptom complexity and frequent overlaps between disease entities, presents particular diagnostic challenges for which language models may be beneficial for clinical decision support (e.g. triage, prioritization), clinical documentation, patient education, literature synthesis, and pattern recognition[4]. Appropriate and reliable decision-making, treatment recommendations, and other guidance provided by such models in routine care are essential for the safety of both patients and physicians. Rheumatologists reported anticipation of positive AI effects and would welcome increased AI implementation and dedicated training program[5]. Recent findings indicate that patients often rate AI-generated responses as highly as physician answers, despite experts identifying significant shortcomings in accuracy[6]. However, the clinical relevance of current-generation LLMs including e. g. GPT-4o, Nemotron (Llama-3.1-70b-instruct), Qwen-Turbo 2.5, and Claude-3.5-Sonnet remains subject to rigorous validations [7,8]. "Small" language models

(SLMs) such as Mixtral-8x7b-32768 are also used in medical scenarios, however, these models also require validation studies[9].

Ongoing concerns regarding language models include hallucination phenomena and their reliance on static or outdated training corpora[11]. Retrieval-augmented generation (RAG) has been proposed to address language models' limitations such as hallucination phenomena and data constraints by enabling langue models to dynamically incorporate current external knowledge such as clinical guidelines during output generation[10,11]. Although RAG has shown early promise in general medical topics, its domain-specific effectiveness, particularly in rheumatology, still requires systematic comparative evaluations[12].

We present a comparative evaluation of five state-of-the-art language models GPT-4o, Mixtral-8x7b-32768, Nemotron (Llama-3.1-70b-instruct), Qwen-Turbo 2.5, and Claude-3.5-Sonnet as applied to rheumatological diagnosis and therapy planning. The models were selected based on their performance in public benchmark platforms, such as LLM Arena[13], with particular emphasis on advanced architectures including Mixture-of-Experts (MoE) systems[14] due to their scalability and efficiency. Representative patient cases were derived from anonymized clinical records from a tertiary rheumatology clinic in Germany. Models were assessed with and without RAG integration and evaluated using the F1 score for diagnostic and therapeutic accuracy, as well as the Retrieval-Augmented Generation Assessment Score (RAGAS) to quantify the factual alignment and relevance of model-generated content. The RAG system employed clinical guideline documents indexed in a FAISS-based vector database[10], with Sentence-BERT for embedding generation and FlashRank for passage re-ranking.

Substantial performance variability was observed across models and configurations (see Table 1, Fig. 1). On average, RAG integration yielded improvements in diagnostic and treatment recommendation accuracy. Mixtral-8x7b-32768 with RAG demonstrated the highest performance in both diagnostic and therapeutic categories when no pre-specified diagnosis was provided (Diagnosis F1: 72%; Treatment F1: 73%), highlighting the effectiveness of its Mixture-of-Experts architecture in leveraging external knowledge[14].

The benefit of RAG integration was found to be model-dependent. Nemotron achieved a high diagnostic F1 score (71%) even without RAG, indicating strong internal knowledge representation. Similarly, Qwen-Turbo performed well for treatment recommendations without

RAG (F1: 72%). Claude-3.5-Sonnet consistently underperformed relative to Mixtral 8x7b and Nemotron-70b-instruct in both categories.

The RAGAS analysis further illuminated model-specific differences in RAG utility (Fig. 1). Mixtral with RAG achieved the highest RAGAS score (81%), reinforcing its superior performance in integrating external data. Notably, Claude-3.5-Sonnet achieved a high RAGAS score without RAG (80%), suggesting a robust baseline performance that is less dependent on RAG integration. Nemotron exhibited the highest performance variability across configurations (range: 51%–67%), indicating greater sensitivity to external knowledge pipelines. GPT-4o showed moderate, consistent performance across all test conditions.

**Table 1 | Summary of Top F1 Scores and RAGAS Scores for LLM Performance Evaluation**

| Model | Model Size | Configuration | Metric | Top Score (%) | Evaluation Assessment |
|---|---|---|---|---|---|
| **Mixtral-8x7b-32768** | SLM | RAG, No Pre-Diagnosis | F1-Dx | 72 | Best diagnostic performance with RAG |
| **Mixtral-8x7b-32768** | SLM | RAG, No Pre-Diagnosis | F1-Tx | 73 | Best treatment recommendation performance |
| **Nemotron 70b instruct** | SLM | No RAG, No Pre-Diagnosis | F1-Dx | 71 | Strong diagnostic performance without RAG |
| **Qwen-Turbo** | SLM | No RAG, No Pre-Diagnosis | F1-Tx | 72 | Strong treatment performance without RAG |
| **GPT-4o** | LLM | RAG, With Pre-Diagnosis | RAGAS | 74 | Good factual alignment with external knowledge |
| **Mixtral-8x7b-32768** | SLM | RAG, With Pre-Diagnosis | RAGAS | 81 | Highest RAGAS score |
| **Claude-3.5-Sonnet** | LLM | No RAG, With Pre-Diagnosis | RAGAS | 80 | High RAGAS score without RAG integration |

*F1-Dx: F1 Score for Diagnosis; F1-Tx: F1 Score for Treatment Recommendation. Scores represent the highest achieved in the respective category across test conditions. Full results are available in Supplementary Information*

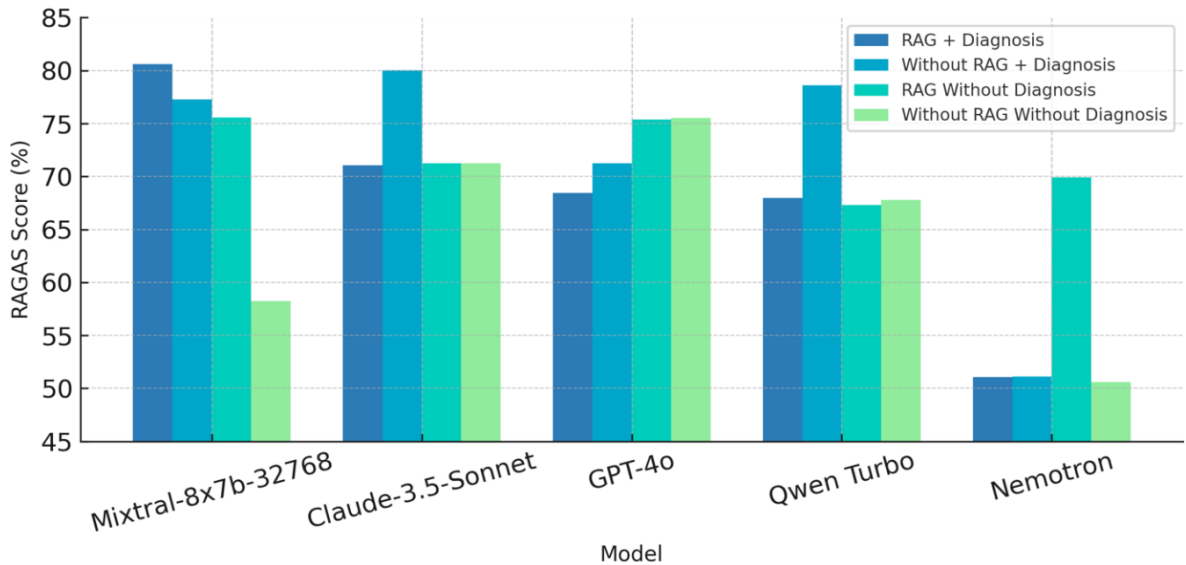

**Figure 1. Types of sequence alignments.** RAGAS Evaluation Scores for Model Response Quality. Bar chart showing RAGAS scores for each model under four conditions: with/without RAG and with/without a pre-defined diagnosis provided. Higher scores indicate better quality of generated medical content according to the RAGAS framework. Mixtral (RAG, with diagnosis) and Claude (No RAG, with diagnosis) show top performance. Nemotron exhibits significant variability.

Prior evidence suggests that LLMs can pass formal medical assessments, such as exam-style questions; however, these results do not necessarily translate into reliable performance in real-world clinical settings[15,16]. Our findings indicate that while LLMs – particularly when combined with RAG - can support diagnosis and treatment planning in rheumatology, their performance remains inconsistent[17]. The effectiveness of RAG depends on the base model's architecture and potentially on its internal knowledge[18]. Discrepancies between quantitative scores and potential clinical applicability highlight the need for (further) careful validations. Although language-model settings such as Mixtral-8x7b-32768+RAG show promise, the occurrence of errors necessitates that artificial intelligence in its current state may serve as a support tool, not as a replacement for clinical expertise. Continuous updating and adaptation to current clinical guidelines and recommendations are crucial[19], technical developments need to be monitored and included in further developments and rheumatology use cases[5,20].

Importantly, minor variations in quantitative metrics such as F1 or RAGAS may mask clinically relevant or potentially harmful errors in model output. From a clinical standpoint, recommendations that are statistically close to correct may still pose substantial risks if, for instance, contraindications, red-flag symptoms, or nuanced contextual factors are overlooked. Such misjudgments, while marginal from a numerical perspective, can lead to serious or even life-threatening consequences in real-world practice[21]. These findings underscore the essential role of expert oversight and the necessity of domain-specific validation.

Although this study did not directly compare LLM or SLM performance with general practitioners, the question of whether such systems reach a "primary care level" is relevant. Some models, particularly Mixtral -8x7b-32768 with RAG, showed performance approaching that of expert systems, yet true equivalence with general practitioners medical reasoning requires an additional handling of diagnostic uncertainty, comorbidities, and patient context factors beyond the scope of this study. Future research should include comparisons with general practitioners and physicians from other domains, such as orthopedics, to better assess real-world applicability.

Beyond supporting individual recommendations, LLMs and SLMs may also accelerate diagnostic timelines by summarizing prior documentation—referral letters, discharge summaries, and lab results—and generating context-specific suggestions. In Germany, the average time from symptom onset to confirmed rheumatoid arthritis diagnosis is approximately 14.2 months[3], contributing to disease progression and higher long-term costs. There is currently only one rheumatologist per 100,000 adults in Germany, while at least two are recommended for adequate care[3]. Waiting times often exceed three months for an initial consultation. In addition to established clinical pathways, such as early arthritis clinics or other fast-track programs, language models could help alleviate bottlenecks by synthesizing existing records and supporting earlier triage and prioritization[22], thereby improving access, reducing the burden on specialists, and enhancing overall resource allocation.

Limitations of this study include the small number of representative test cases and reliance on specific metrics (F1, RAGAS). Although prompt engineering was standardized, further model-specific tuning could yield different results[23,24]. All results were benchmarked against evidence-based guideline recommendations.

In conclusion, LLMs and SLMs hold promise for augmenting clinical decision-making in rheumatology, particularly when paired with retrieval-augmented generation. SLMs such as Mixtral-8x7b-32768, when properly configured, can achieve high performance while offering practical advantages for clinical deployment. Importantly, their lower computational footprint, reduced cost, and deployability in low-resource settings make them strong candidates for economically sustainable integration into healthcare workflows. Nonetheless, current systems do not yet reach specialist-level reliability and must not be used without expert oversight. Future efforts should prioritize clinical validation at scale and the development of evaluation frameworks aligned with both medical and economic standards.

## Methods

**Study Design and Data.** An experimental approach was used, based on primary data analysis of anonymized medical records from rheumatology patients in a tertiary center. Ten standardized patient cases were created, containing information on diagnoses, medications, laboratory values, and therapy progress. Documents were pre-processed for uniform structure.

**Models Tested.** Five language models were evaluated: ChatGPT-4o (OpenAI), Mixtral-8x7b-32768 (Mistral AI), Llama-3.1-nemotron-70b-instruct (Nvidia, referred to as Nemotron), Qwen-Turbo 2.5 (Alibaba Cloud), and Claude-3.5-Sonnet-20240620 (Anthropic).

**Retrieval-Augmented Generation (RAG) Implementation.** The RAG system involved indexing relevant medical guidelines (e.g., EULAR, ACR, SIGN) using a FAISS (IndexFlatL2) vector store[25]. Text segments were converted to 768-dimensional embeddings using Sentence-BERT ('distilbert-base-nli-stsb-mean-tokens'). Retrieved passages were re-ranked using FlashRank before being provided as context to the LLMs.

**Evaluation Procedure.** Each model was tested under four conditions: with RAG / without RAG, and with or without a pre-defined diagnosis provided. Standardized prompts were used across models, incorporating techniques like few-shot learning and chain-of-thought prompting[23,24]. Model outputs (diagnosis and treatment recommendations) were evaluated against guideline-based ground truths.

**Evaluation Metrics.** Model performance was quantified using:
**F1 Score**: Calculated for both diagnostic accuracy and treatment recommendations, measuring the harmonic mean of precision and recall[26].
**RAGAS (Retrieval-Augmented Generation Assessment Score):** Used to evaluate the quality aspects of responses generated with RAG, including faithfulness and relevance [19].

**Acknowledgments**

N/A

**Data Availability Statement**

Data is available upon reasonable request.

**Funding**

This project was not funded.

**Author information / Authors and Affiliations**

Institute of IT Management and Digitization Research (IFID), FOM University of Applied Sciences, 40476 Düsseldorf, Germany

**Sabine Felde**, **Rüdiger Buchkremer**

Department of Rheumatology, University Hospital Düsseldorf, Heinrich-Heine-University, Düsseldorf, Germany

**Gamal Chehab**, **Jörg HW Distler**, **Matthias Schneider**, **Jutta G. Richter**

Hiller Research Center, Heinrich-Heine-University, Düsseldorf, Germany

**Gamal Chehab, Jörg HW Distler**, **Matthias Schneider**, **Jutta G. Richter**

FOM University of Applied Sciences, Competence Center for Medical Economics, Essen, Germany

**Christian Thielscher**

Department of Rheumatology and Clinical Immunology, KEM Kliniken Essen-Mitte, Essen, Germany

**Gamal Chehab**

Corresponding author

Prof. Dr. Jutta G. Richter
Department for Rheumatology, University Hospital Düsseldorf, Medical Faculty of Heinrich-Heine-University
Hiller Research Center, University Hospital Düsseldorf, Medical Faculty of Heinrich-Heine-University
Moorenstr. 5
40225 Duesseldorf
Germany
Email jutta.richter@med.uni-duesseldorf.de
Phone   +49 211 8117817
FAX     +49 211 8116455

**Author contributions**

The authors of this paper contributed the following:

R. B. conceived the idea of combining small language models with RAGs for clinical decision support. S. F. and R. B. conceived the technical setup including Flash RAG selection. S. F., R. B., C. T., G. C. and J. G. R. designed the initial concept for analyzing anonymized patient data in rheumatoid arthritis. G. C. and J.G.R evaluated the models output. S. F. authored her graduate thesis on this topic. R. B. and C. T. served as S.F.´s graduate advisors on her graduate thesis at the FOM University of Applied Sciences. All authors reviewed and approved the final manuscript.

**Competing interests**

The authors declare that they have no competing interests

**Open Access** This article is licensed under a Creative Commons Attribution 4.0 International License.